\documentclass[12pt]{article}
\pdfoutput=1
\usepackage{fullpage}
\usepackage{natbib}

\usepackage[utf8]{inputenc} 
\usepackage[T1]{fontenc}    
\usepackage{hyperref}       
\usepackage{url}            
\usepackage{booktabs}       
\usepackage{amsfonts}       
\usepackage{nicefrac}       
\usepackage{microtype}      

\usepackage{graphicx}
\usepackage{amsmath}
\usepackage{amssymb}
\usepackage{enumerate}
\usepackage{float}
\usepackage{enumitem}
\usepackage{subfigure}
\usepackage{multirow}
\usepackage{caption}
\usepackage{xcolor}

\renewcommand{\(}{\left(}
\renewcommand{\)}{\right)}

\newcommand{\x}[1]{\mathbf{x}_{#1}}
\newcommand{\z}[1]{\mathbf{z}_{#1}}

\newcommand{\y}{\mathbf{y}}

\newcommand{\p}{p_\theta}
\newcommand{\q}{q_\phi}
\newcommand{\fp}{f_\theta}
\newcommand{\fq}{f_\phi}

\newcommand{\comment}[1]{}

\newcommand{\E}{\mathbb{E}}
\renewcommand{\L}{\mathcal{L}}
\newcommand{\N}{\mathcal{N}}
\renewcommand{\~}{\ \raise.17ex\hbox{$\scriptstyle\mathtt{\sim}$}\ }
\newcommand{\newtilda}{\raise.17ex\hbox{$\scriptstyle\mathtt{\sim}$}}

\newcommand{\footremember}[2]{%
    \footnote{#2}%
    \newcounter{#1}%
    \setcounter{#1}{\value{footnote}}
}
\newcommand{\footrecall}[1]{%
    \footnotemark[\value{#1}]
}

\title{\bf{\LARGE Dr.VAE: Drug Response Variational Autoencoder}}
\author{
  Ladislav Rampasek\footremember{corr}{Corresponding authors: \texttt{rampasek@cs.toronto.edu} and \texttt{anna.goldenberg@utoronto.ca}}\footremember{DCS}{University of Toronto, Department of Computer Science, Toronto, ON, Canada}\footremember{SK}{The Hospital for Sick Children, Toronto, ON, Canada}
  \and Daniel Hidru\footrecall{DCS}\footrecall{SK}
  \and Petr Smirnov\footremember{PMH}{Princess Margaret Cancer Centre, University Health Network, Toronto, ON Canada}
  \and Benjamin Haibe-Kains\footrecall{PMH}\footremember{DMB}{Department of Medical Biophysics, University of Toronto, Toronto, ON, Canada}\footrecall{DCS}\footremember{OICR}{Ontario Institute of Cancer Research, Toronto, ON, Canada}
  \and Anna Goldenberg\footrecall{corr}\footrecall{SK}\footrecall{DCS}
}
\date{}

\begin{document}

\maketitle

\begin{abstract}
We present two deep generative models based on Variational Autoencoders to improve the accuracy of drug response prediction. Our models, Perturbation Variational Autoencoder and its semi-supervised extension, Drug Response Variational Autoencoder (Dr.VAE), learn latent representation of the underlying gene states before and after drug application that depend on: (i) drug-induced biological change of each gene and (ii) overall treatment response outcome. Our VAE-based models outperform the current published benchmarks in the field by anywhere from $3$ to $11\%$ AUROC and $2$ to $30\%$ AUPR. In addition, we found that better reconstruction accuracy does not necessarily lead to improvement in classification accuracy and that jointly trained models perform better than models that minimize reconstruction error independently.
\end{abstract}

\section{Introduction}
Despite tremendous advances in the pharmaceutical industry, many patients worldwide do not respond to the first medication they are prescribed.  \comment{Worse, many have severe adverse reactions to their medications resulting in complications or even death.}  Personalized medicine, an approach that uses patients' own genomic data, promises to tailor the treatment program to increase the probability of positive response. That idea is great, but to build powerful predictive models, we need training data. The space of all possible treatments and their combinations for a given condition is enormous and the heterogeneity of patients with complex diseases is high. Thus, while much data has been collected, it is sparsely and inefficiently sampled making it a very hard learning problem.  

In the last decade there have been several public releases of high throughput drug screening in cell lines. Cancer cell lines are cells taken from a patient's tumor that are ``immortalized'', i.e. modified to divide indefinitely. The greatest advantage of cell lines is that it is relatively inexpensive to test them with thousands of drugs providing a rich basis for learning predictive models. This screening task was undertaken by several large consortia and pharmaceutical companies resulting in public datasets of various sizes, e.g. Genomics of Drug Sensitivity in Cancer (GDSC) with 138 drugs \citep{yang2013} tested on 700 cancer cell lines, and the Cancer Cell Line Encyclopedia (CCLE) \citep{Barretina:2012} with 24 drugs tested on a panel of >1000 cell lines. The availability of these datasets spurred the development of predictive models. \citet{Jang:2014} compared seven standard machine learning approaches and identified ridge and elastic net regressions as the best performers with an average AUC of $\newtilda 0.79$ across 24 compounds from the CCLE dataset and $\newtilda 0.75$ across 138 compounds from the GDSC dataset. \comment{Since the number of cell lines that have been tested with the same drug is relatively small ($\newtilda 500$) compared to the number of genes in the dataset ($ \newtilda 12,000$), multitask methods were shown to perform better in some cases. For example, Kernelized Bayesian multitask model \citep{gonen:2014} not substantially, but consistently outperformed single-task models in predicting drug response for most of the 138 drugs.}

In addition, there is a perturbation database containing over 16,000 experiments showing how the expression of 1000 genes changed in response to a drug (gene expression is recorded before and after drug application) \citep{Duan:2014}. This information allows for the assessment of biological change due to treating the cancer cells with drugs. Combining response and perturbation data is expected to ultimately yield a better and more biologically relevant model of drug response, though likely more experiments will be needed, since there are only a few drugs tested in each cell line. 

\comment{Recently introduced expressive generative models like Generative Adversarial Networks (GAN) \citep{goodfellow2014GAN} and Variational Autoencoders (VAE) \citep{kingma2013VAE, rezende2014VAE} became popular for modeling various generative tasks (image generation, style transfer, super-resolution generation), active learning, etc. Recently \citet{kadurin2017} used an Adversarial Autoencoder \citep{makhzani2015AAE} as a proof-of-concept for drug discovery. While GANs tend to fair better in the generative task itself,} 

In this paper we present two deep generative models Perturbation Variational Autoencoder and its semi-supervised extension,  Drug Response Variational Autoencoder (Dr.VAE), that learn latent representation of the underlying gene states before and after drug application that depend on both the cell line's overall response to the drug and the biological change of each of the landmark genes. We are building on VAEs ability to leverage expressiveness of deep neural networks for Bayesian learning and inference \citep{kingma2014SSVAE}. In addition, as Bayesian models they are more adept for the task when very little data is present, which is the case in our drug response prediction problem. To fit our model we use a combination of Stochastic Gradient Variational Bayes \citep{kingma2013VAE} and Inverse Autoregressive Flow \citep{kingma2016IAF}, a recently introduced type of Normalizing Flow \citep{rezende2015NF}.

We tested our methods on 19 drugs for which both perturbation and drug response data was available. Both Dr.VAE and Semi-Supervised VAE outperform the current published benchmark models \citep{Jang:2014} in the field by anywhere from $3$ to $11\%$ AUROC and $2$ to $30\%$ AUPR.
\comment{Our experiments show that adding unlabeled data helps the performance, including perturbation experiments also improves the performance, though interestingly, better predictive models have higher reconstruction error, compared to models where reconstruction was optimized. Additionally, PCA-type reconstruction seems to do well, but using the embedded space for prediction does not do quite as well as using our Perturbation VAE. Overall, this means that the boost in prediction performance comes primarily from capturing more of the noise in the gene expression measurements compared to faithful latent embedding.}
Our analysis of this problem and of the model performance shows the difficulty of fitting sparsely and inefficiently sampled high dimensional data, but at the same time illustrates the flexibility and potential improvement over the currently available state-of-the-art models for drug response prediction problem.

\section{Methods}
We propose two models. First, we discuss an approach for modeling drug perturbation effects, i.e. given gene expression of a cell line before the drug is applied (pre-treatment gene expression), we are aiming to predict gene expression after the drug is applied (post-treatment state). To this end we propose a deep generative model, Perturbation VAE (PertVAE).

We then develop drug response prediction model, Drug Response Variational Autoencoder (Dr.VAE), a semi-supervised extension of PertVAE, to tackle the problem of drug response (treatment efficacy) prediction while harnessing the unsupervised information about the particular drug from observed pre- and post-treatment gene expression perturbation pairs.

\subsection{Perturbation VAE}
Perturbation Variational Autoencoder (PertVAE) is an unsupervised model for drug-induced gene expression perturbations, that embeds the data space (gene expression) in a lower dimensional latent space. In the latent space we model the drug-induced effect as a linear function, which is trained jointly with the embedding encoder and decoder.

\begin{figure}
\centering
\subfigure[]{ 
	\begin{minipage}[]{0.24\textwidth}
	\centering
	\includegraphics[width=1\textwidth]{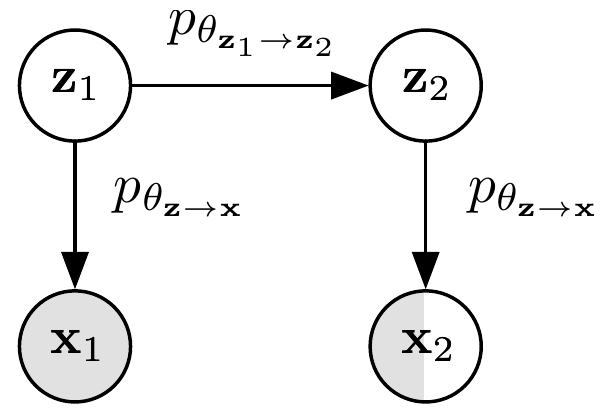}
		\end{minipage}
}
\hspace{25pt}
\subfigure[]{ 
	\begin{minipage}[]{0.24\textwidth}
	\centering
	\includegraphics[width=1\textwidth]{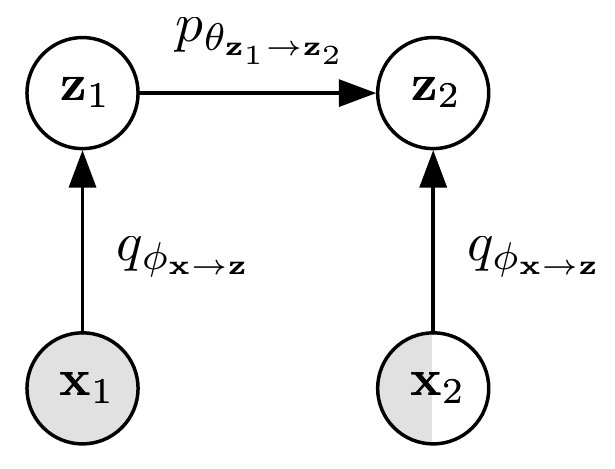}
	\end{minipage}	
}
\caption{Perturbation VAE: (a) Factorization of the generative distribution $p$, (b) Factorization of the approximate posterior distribution $q$. Note, we use the generative ${\p}_{\z{1}\rightarrow\z{2}}$ in case $\x{2}$ is not observed.}\label{fig:pertvae_model}
\end{figure}

We fit PertVAE on ``perturbation pairs'' $[\x{1}, \x{2}]$ of pre-treatment and post-treatment gene expression with shared stochastic embedding encoder ${\q}_{\x{}\rightarrow\z{}}$ and decoder ${\p}_{\z{}\rightarrow\x{}}$. The original dimension of each vector $\x{}$ is $903$ genes. Additionally we use unpaired pre-treatment data (with no know post-treatment state) to improve learning of the latent representation. The graphical representation of PertVAE model is shown in Figure \ref{fig:pertvae_model}.

\paragraph{Joint distribution.} Our Perturbation VAE models joint $p(\x{1}, \x{2}, \z{1}, \z{2})$, which we assume to factorize as:
\begin{align}
p(\x{1}, \x{2}, \z{1}, \z{2}) = p(\x{1} | \z{1}) \cdot p(\x{2} | \z{2}) \cdot p(\z{2} | \z{1}) \cdot p(\z{1}) 
\end{align}

\paragraph{Generative distribution $p$.} Perturbation VAE's generative process is as follows:
\begin{align}
p(\z{1}) &= \N\(\mathbf{0}, \mathbf{I}\) \\
{\p}_{\z{1}\rightarrow\z{2}}(\z{2} | \z{1}) &= \N\(\z{2} | \boldsymbol{\mu}_{\z{2}} = \fp(\z{1}), \boldsymbol{\sigma}_{\z{2}} = \exp^{\fp(\z{1})} \) \\
k\in\{1, 2\}: {\p}_{\z{}\rightarrow\x{}}(\x{k} | \z{k}) &= \N\(\x{k} | \boldsymbol{\mu}_{\x{k}} = \fp(\z{k}), \boldsymbol{\sigma}_{\x{k}} = \exp^{\fp(\z{k})} \)
\end{align}

The parameters of these distributions are computed by functions $\fp{}$, which are neural networks with a total set of parameters $\Theta$. For brevity we refer to these parameters as $\Theta$ instead of more specific subsets $\theta_{\z{}\rightarrow\x{}}$ or $\theta_{\z{1}\rightarrow\z{2}}$ when such level of detail unnecessarily clutters the notation.

We constrain the mean function in ${\p}_{\z{1}\rightarrow\z{2}}$ to be a linear function ${\fp}_{\z{1}\rightarrow\z{2}}(\z{1})$ of the following form:
\begin{align}
{\fp}_{\z{1}\rightarrow\z{2}}(\z{1}) \equiv \z{1} + \mathbf{W}\z{1} + \mathbf{b}
\end{align}
with $\mathbf{W}$ and $\mathbf{b}$ initialized close to zero such that ${\fp}_{\z{1}\rightarrow\z{2}}(\z{1})$ starts as an identity function. We found that together with L2 penalization this formulation improves stability and generalization of the model.

\paragraph{Approximate posterior $q$.} Depending on the type of the data, we assume the approximate posterior $q$ with a set of parameters $\phi$ to factorize as:
\begin{align}
\text{perturbation pairs: \ \ }& \q(\z{1},\z{2} | \x{1},\x{2}) = {\q}_{\x{}\rightarrow\z{}}(\z{1} | \x{1}) \cdot {\q}_{\x{}\rightarrow\z{}}(\z{2} | \x{2}) \\
\text{pre-treatment singleton: \ \ }& \q(\z{1},\z{2},\x{2} | \x{1}) = {\q}_{\x{}\rightarrow\z{}}(\z{1} | \x{1}) \cdot {\p}_{\z{1}\rightarrow\z{2}}(\z{2} | \z{1}) \cdot {\p}_{\z{}\rightarrow\x{}}(\x{2} | \z{2})
\end{align}
Analogously to the shared generative ${\p}_{\z{}\rightarrow\x{}}$ distribution, also ${\q}_{\x{}\rightarrow\z{}}(\z{k} | \x{k})$ is shared for both $k\in\{1, 2\}$. Here, instead of directly using a diagonal Gaussian as the final approximate posterior
\begin{align}
k\in\{1, 2\}: {\q}_{\x{}\rightarrow\z{}}(\z{k} | \x{k}) &= \N\(\z{k} | \boldsymbol{\mu}_{\z{k}} = \fq(\x{k}), \boldsymbol{\sigma}_{\z{k}} = \exp^{\fq(\x{k})} \)
\end{align}
we apply two steps of ``LSTM-type'' Inverse Autoregressive Flow (IAF) \citep{kingma2016IAF} updates to facilitate a richer family of approximate distributions. A sample from ${\q}_{\x{}\rightarrow\z{}}(\z{k} | \x{k})$ is then derived by two iterations of the following step:
\begin{align}
\z{k}^{(0)} &\sim \N\(\boldsymbol{\mu}_{\z{k}} = \fq(\x{k}), \boldsymbol{\sigma}_{\z{k}} = \exp^{\fq(\x{k})} \) \\
t\in\{1, 2\}: \z{k}^{(t)} &= \text{sigmoid}(\mathbf{s}^{(t)}) \odot \z{k}^{(t-1)} + (1-\text{sigmoid}(\mathbf{s}^{(t)})) \odot \mathbf{m}^{(t)}
\end{align}
The coefficients $[\mathbf{m}^{(t)}, \mathbf{s}^{(t)}]$ of the IAF are computed by a Masked Autoencoder for Distribution Estimation (MADE) model \citep{germain2015made}:
\begin{align}
[\mathbf{m}^{(t)}, \mathbf{s}^{(t)}]  = \text{MADE}^{(t)}\(\z{k}^{(t-1)}, \mathbf{h}(\x{k})\)
\end{align}
MADE is an autoregressive model, that is, $j$-th elements of the $\mathbf{m}^{(t)}$ and $\mathbf{s}^{(t)}$ vectors only depend on up to the first $j-1$ elements of $\z{k}^{(t-1)}$. Using this property, the determinant of Jacobian of each IAF step can be computed efficiently. As each IAF step is then an invertible function with known Jacobian determinant, it is thus possible to efficiently compute probability of the derived sample $\z{k}^{(t)}$ in the complex posterior ${\q}_{\x{}\rightarrow\z{}}(\z{k} | \x{k})$ that does not have a parametric form \citep{kingma2016IAF}.

\paragraph{Fitting $\theta$ and $\phi$ parameters.}
We jointly optimize the generative model $\theta$ and variational $\phi$ parameters with Stochastic Gradient Variational Bayes (SGVB) \citep{kingma2013VAE,rezende2014VAE} to maximize Evidence Lower Bound (ELBO) of the data:
\begin{align}
\sum^{N_{P}} \log p\(\x{1}, \x{2}\) + \sum^{N_{S}} \log p\(\x{1}\) \geq \text{ELBO}_{\text{PertVAE}} \\
\text{ELBO}_{\text{PertVAE}} = \sum^{N_{P}} \L_{P}\(\x{1}, \x{2};\theta, \phi\) + \sum^{N_{S}} \L_{S}\(\x{1};\theta, \phi\)
\end{align}
which is a sum of the evidence lower bound of $N_{P}$ perturbation pairs and the lower bound of $N_{S}$ unpaired ``singleton'' examples that we leverage to train the latent space Variational Autoencoder as well. The individual per-example lower bounds $\L_{P}$ and $\L_{S}$ take the following form:
\begin{align}
	\L_{P}(\x{1}, \x{2};\ \theta, \phi)
&=	\E_{\q(\z{1},\z{2} | \x{1},\x{2})} \big[
		\log \p(\x{1},\x{2},\z{1},\z{2}) - \log \q(\z{1},\z{2} | \x{1},\x{2}) \big]\\
&=	\E_{\q(\z{1} | \x{1})}\left[\log\p(\x{1} | \z{1}) \right] -D_{KL}\left[\q(\z{1} | \x{1}) || p(\z{1}) \right] +
\\&\phantom{=\ } + \E_{\q(\z{2} | \x{2})}\left[\log\p(\x{2} | \z{2}) \right] -D_{KL}\left[\q(\z{2} | \x{2}) || \p(\z{2} | \z{1}) \right] \nonumber \\
	\L_{S}(\x{1};\ \theta, \phi)
&=	\E_{\q(\z{1} | \x{1})} \big[ \log \p(\x{1},\z{1}) - \log \q(\z{1} | \x{1}) \big]\\
&=	\E_{\q(\z{1} | \x{1})}\left[\log\p(\x{1} | \z{1}) \right] -D_{KL}\left[\q(\z{1} | \x{1}) || p(\z{1}) \right] \nonumber
\end{align}
Using SGVB it is possible to backpropagate through $\text{ELBO}_{\text{PertVAE}}$ and we use Adam \citep{kingma2014adam} to compute gradient updates for both $\theta$ and $\phi$ parameters. As we use IAF to model $\q(\z{k} | \x{k})$, the Kullback–Leibler divergence $D_{KL}$ cannot be computed numerically and therefore we use a Monte Carlo estimate. Additionally we follow \citep{kingma2016IAF} and allow ``free bits'' in $D_{KL}$ to mitigate the problem of overly strong prior causing the optimization to get stuck in bad local optima.

\subsection{Drug Response VAE}
Analogously to Semi-Supervised VAE, we can extend our unsupervised Perturbation VAE to a semi-supervised model by stacking a modified ``M2 model'' \citep{kingma2014SSVAE}. This model can be trained jointly to model both drug-induced perturbation effects as well as treatment response outcomes. As such we call this model Drug Response VAE (Dr.VAE).

\begin{figure}
\centering
\subfigure[]{ 
	\begin{minipage}[]{0.19\textwidth}
	\centering
	\includegraphics[width=1\textwidth]{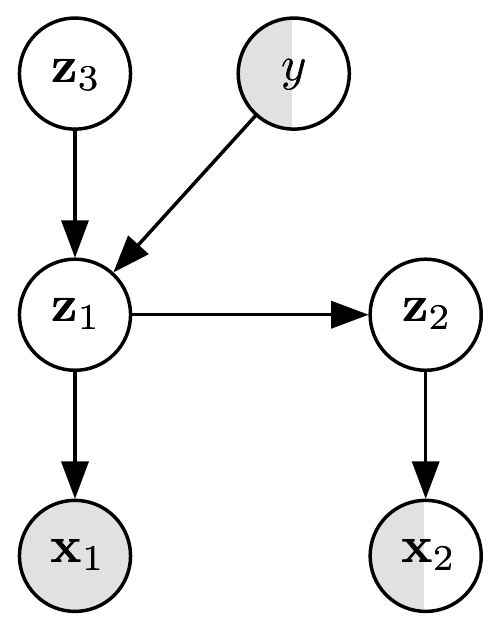}
		\end{minipage}
}
\hspace{25pt}
\subfigure[]{ 
	\begin{minipage}[]{0.19\textwidth}
	\centering
	\includegraphics[width=1\textwidth]{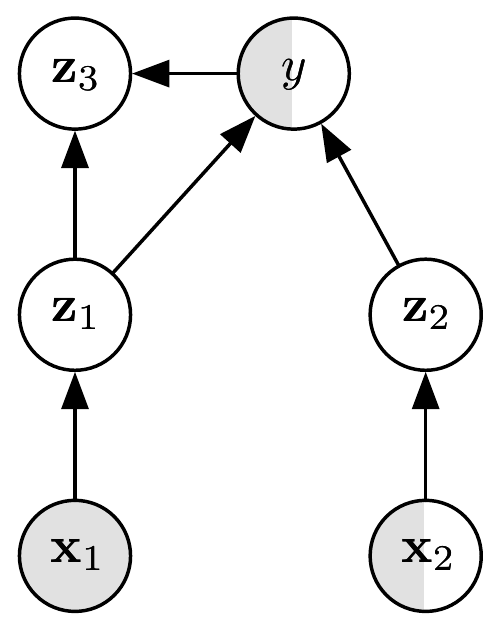}
	\end{minipage}	
}
\caption{Dr.VAE is semi-supervised extension of our Perturbation VAE: (a) Factorization of the generative distribution $p$, (b) Factorization of the approximate posterior distribution $q$. In case the post-treatment gene expression $\x{2}$ is not observed, we can use the expected posterior $\E_{\q(\z{1} | \x{1})}\left[\p(\z{2} | \z{1}) \right]$ for $\z{2}$ instead.}\label{fig:drvae_model}
\end{figure}

We use similar type of data to train Dr.VAE as we use for PertVAE, however some of the perturbation pairs and pre-treatment singletons now can have a binary outcome label $y$ associated with them, denoting if the drug treatment was successful or not. Schema of Dr.VAE model is shown in Figure \ref{fig:drvae_model}.

\paragraph{Joint distribution.} Drug Response VAE extends Perturbation VAE to model a joint distribution $p(\x{1}, \x{2}, \z{1}, \z{2}, \z{3}, \y)$ factorized as:
\begin{align}
p(\x{1}, \x{2}, \z{1}, \z{2}, \z{3}, \y) = p(\x{1} | \z{1}) \cdot p(\x{2} | \z{2}) \cdot p(\z{2} | \z{1}) \cdot p(\z{1} | \z{3}, \y) \cdot p(\z{3}) \cdot p(\y)
\end{align}

\paragraph{Generative distributions $p$.} The individual generative distributions Dr.VAE factorizes have the following form:
\begin{align}
p(\y) &= \text{Cat}(\y) \\
p(\z{3}) &= \N\(\mathbf{0}, \mathbf{I}\) \\
\p(\z{1} | \z{3}, \y) &= \N\(\z{1} | \boldsymbol{\mu}_{\z{1}} = \fp(\z{3}, \y), \boldsymbol{\sigma}_{\z{1}} = \exp^{\fp(\z{3}, \y)} \) \\
\p(\z{2} | \z{1}) &= \N\(\z{2} | \boldsymbol{\mu}_{\z{2}} = \fp(\z{1}), \boldsymbol{\sigma}_{\z{2}} = \exp^{\fp(\z{1})} \) \\
k\in\{1, 2\}: \p(\x{k} | \z{k}) &= \N\(\x{k} | \boldsymbol{\mu}_{\x{k}} = \fp(\z{k}), \boldsymbol{\sigma}_{\x{k}} = \exp^{\fp(\z{k})} \)
\end{align}
Same way as in PertVAE, we share the ``data decoder'' $\p(\x{k} | \z{k})$ among both $k\in\{1, 2\}$.

\paragraph{Approximate posterior $q$.} Depending on the type of the data, we assume the approximate posterior $q$ to factorize as:
\begin{align}
\text{\small labeled pair:\ }& \q(\z{1},\z{2},\z{3} | \x{1},\x{2},\y) = \q(\z{1} | \x{1}) \cdot \q(\z{2} | \x{2}) \cdot \q(\z{3} | \z{1}, \y) \\
\text{\small unlabeled pair:\ }& \q(\z{1},\z{2},\z{3},\y | \x{1},\x{2}) = \q(\z{1} | \x{1}) \cdot \q(\z{2} | \x{2}) \cdot \q(\y | \z{1}, \z{2}) \cdot \q(\z{3} | \z{1}, \y) \\
\text{\small labeled singleton:\ }& \q(\z{1},\z{2},\z{3},\x{2} | \x{1},\y) = \q(\z{1} | \x{1}) \cdot \p(\z{2} | \z{1}) \cdot \p(\x{2} | \z{2}) \cdot \q(\z{3} | \z{1}, \y) \\
\text{\small unlab. singleton:\ }& \q(\z{1},\z{2},\z{3},\x{2},\y | \x{1}) = \q(\z{1} | \x{1}) \cdot \p(\z{2} | \z{1}) \cdot \p(\x{2} | \z{2}) \cdot 
\\&\phantom{\q(\z{1},\z{2},\z{3},\y | \x{1},\x{2}) =\ } \cdot \q(\y | \z{1}, \z{2}) \cdot \q(\z{3} | \z{1}, \y) \nonumber
\end{align}
The ``data encoder'' $k\in\{1, 2\}: \q(\z{k} | \x{k})$ is shared and parametrized the same way as in PertVAE. The additional approximate posterior distributions then take the following form:
\begin{align}
\q(\y | \z{1}, \z{2}) &= \text{Cat}\big(\y | \boldsymbol{\pi}=\text{softmax}(\fq(\z{1}, \z{2})) \big) \\
\q(\z{3} | \z{1}, \y) &= \N\(\z{3} | \boldsymbol{\mu}_{\z{3}} = \fq(\z{1}, \y), \boldsymbol{\sigma}_{\z{3}} = \exp^{\fq(\z{1}, \y)} \) 
\end{align}

The afford mentioned factorizations of the joint and of the posteriors also provide a recipe for sampling and inference in the model by Monte Carlo sampling.

\paragraph{Fitting $\theta$ and $\phi$ parameters.}
We have 4 different sets of partially observed variables, which correspond to different types of data. Therefore there are 4 different evidence lower bounds to optimize:
\begin{align}
N_{LP} \text{ labeled perturbation pairs: \ }& \textstyle \sum^{N_{LP}} \L_{LP}\(\x{1}, \x{2}, \y;\theta, \phi\) \\
N_{UP} \text{ unlabeled perturbation pairs: \ }& \textstyle \sum^{N_{UP}} \L_{UP}\(\x{1}, \x{2};\theta, \phi\) \\
N_{LS} \text{ labeled pre-treatment singletons: \ }& \textstyle \sum^{N_{LS}} \L_{LS}\(\x{1}, \y;\theta, \phi\) \\
N_{US} \text{ unlabeled pre-treatment singletons: \ }& \textstyle \sum^{N_{US}} \L_{US}\(\x{1};\theta, \phi\)
\end{align}
The sum of these 4 specific evidence lower bounds, $\text{ELBO}_{\text{DrVAE}}$, is the evidence lower bound we need to maximize. We omit the derivation of these specific lower bounds in the main manuscript since it follows the same principles as shown above for PertVAE and as shown in the derivation of Semi-Supervised VAE \citep{kingma2014SSVAE, louizos2015VFAE}.

Finally, we need to explicitly introduce loss of the predictive posterior $\log\q(\y | \z{1}, \z{2})$ in order for it to be trained also on labeled data. This is required as for the labeled data the random variable $\y$ is observed and therefore the lower bounds $\L_{LP}$ and $\L_{LS}$ are conditioned on $\y$ and do not contribute to fitting of $\q(\y | \z{1}, \z{2})$. Our final objective $\mathcal{J}_{\text{DrVAE}}$ to maximize is
\begin{align}
\begin{split}
\mathcal{J}_{\text{DrVAE}} &= \text{ELBO}_{\text{DrVAE}} + \alpha \sum^{N_{LP}} \E_{\q(\z{1},\z{2} | \x{1}, \x{2})}\left[ - \log\q(\y = \mathbf{t} | \z{1}, \z{2}) \right] + \\
&\phantom{=\ } + \alpha \sum^{N_{LS}} \E_{\q(\z{1},\z{2} | \x{1})}\left[ - \log\q(\y = \mathbf{t} | \z{1}, \z{2}) \right]
\end{split}
\end{align}

\section{Datasets}
In our experiments we used two main data resources: (i) high-throughput screens of cancer cell-lines including gene expression pre-treatment for all tested cell lines and drug response in terms of cell line viability, and (ii) high-throughput screens of gene expression perturbation effects induced by drugs in cancer cell lines.

We tested our methods on a panel of 19 drugs for which there are both response and perturbation experiments available. These 19 drugs were also used in recent AstraZeneca-Sanger DREAM Challenge and therefore we use it as a representative sample of anti-cancer drugs.

\subsection{Datasets of drug response in cancer cell lines}
Large high-throughput screening efforts have been undertaken resulting in publicly available datasets of cancer cell lines with post treatment cell viability at various drug concentrations. In our experiments we utilize the Genomics of Drug Sensitivity in Cancer (GDSC) \citep{yang2013} and Cancer Cell Line Encyclopedia (CCLE) \citep{Barretina:2012} datasets. We obtained these datasets using PharmacoGx R package \citep{smirnov2015}. As the drug response outcome we use binarized version of dose-response curves, which were computed by PharmacoGx. For consistency, we use response outcome from GDSC, while we use all the other cell lines in GDSC and CCLE not screened for response to a given drug as unlabeled cell line examples. Summary of our pooled dataset is detailed in Supplementary Materials.

\subsection{Drug-induced perturbations dataset}
The Library of Network-Based Cellular Signatures (LINCS) consortium screened perturbation effects that drugs have on gene expression of L1000 landmark genes in cancer cell lines \citep{Duan:2014}. These experiments do not measure the drug treatment efficacy, however some of these cell lines were independently tested in GDSC for the drug response. We cross-reference these cell lines and assign the corresponding label to their perturbation measurement. 

The L1000 perturbation dataset is very sparse: for the 19 drugs, only up to 56 different cell lines were screened. In fact, only 8 drugs have been measured in over 50 cell lines, the remaining 19 have been measured in fewer than 20 cell lines, albeit at various concentrations and with many biological replicates. In our results we use measurements at the highest drug concentration and all the biological replicates of such experiments. In cross-validation of our models we use cell-line-wise splitting so that the biological replicates for a particular cell line are in the same data fold.


\section{Results}
We tested the performance of our models on three tasks: (i) drug response prediction task, (ii) drug perturbation prediction and (iii) gene expression reconstruction from the latent representation.

\paragraph{Architecture.} All evaluated Variational Autoencoder -based models, our proposed models (Dr.VAE, PertVAE) and the published models we used for comparison (VAE and Semi-Supervised VAE (SSVAE)), use 100 stochastic latent units, i.e. all $\z{}$ are stochastic vectors of length 100, and have the same architecture for the ``data encoder'' $\q(\z{k}|\x{k})$ and ``data decoder'' $\p(\x{k}|\z{k})$. For the encoder, there are 903 input units corresponding to 903 landmark genes (we exclude $\newtilda 70$ genes that we could not uniquely map between data sets). The encoder has hidden layers with 500 and 300 units from which parameters of initial Gaussian distribution $\boldsymbol{\mu}_{\z{k}}$ and $\boldsymbol{\sigma}_{\z{k}}$ are computed together with 200 hidden units on which the subsequent Inverse Autoregressive Flow is conditioned. We use 2 steps of IAF, each with one hidden layer of 300 units. Architecture of data decoder mirrors that of data encoder, but without IAF. Throughout all our models we use ELU activation function \citep{clevert2015elu} and Weight Normalization \citep{salimans2016wnorm}. We preserve various parts of the architecture among different models to help with further analysis of what helps with the observed performance.

For both Dr.VAE and SSVAE, the classification posteriors $\log\q(\y | \z{1}, \z{2})$ and $\q(\y | \z{1})$, respectively, are linear functions with soft-max activation over two output units. In our implementation, we use a slight modification for Dr.VAE, for which we found that using $[\z{1}, \z{2}-\z{1}]$ instead of $[\z{1},\z{2}]$ as the classifier input improves the performance. Further, the distributions $\p(\z{1} | \z{3}, \y)$ and $\q(\z{3} | \z{1}, \y)$ (and their equivalents in SSVAE) are parametrized by a neural network with a single hidden layer of 100 units.

All our presented experiments are evaluated in 10-times randomized 5-fold cross-validation and we report the average metric across these 50 data splits. The models were fitted independently for each of the 19 drugs, but with the same hyperparameters.

\subsection{Predicting drug response}
We compare our models to Ridge L2 logistic regression (LR), random forest (RF), and support vector machine with a linear kernel (SVM), following \citet{Jang:2014} that found Ridge LR to be the best overall classifier for drug response in GDSC dataset. 

To assess informativeness of drug-induced perturbations for drug response prediction task we also  compare Dr.VAE to a Semi-Supervised VAE \citep{kingma2014SSVAE}. SSVAE is trained on the same data, however without ability to model the drug effect, as the perturbation pairs are simply presented as  independent unlabeled singletons. On average, Dr.VAE is the best performing method from all tested models ranging from $1\%$ to $30\%$ improvement over the ridge logistic regression, considered state-of-the-art in the field. Over all 19 drugs, the average improvement in performance is $8.95\%$ for Dr.VAE compared to $8.07\%$ of SSVAE. The only drug where both Dr.VAE and SSVAE performed worse than Ridge LR is paclitaxel. This is a chemotherapy drug (no specific gene target) with a much smaller sample size, thus it appears that the simpler model has an advantage over all other models for this one. 

Dr.VAE and SSVAE learn a latent representation of the data and the classifier jointly. To understand the importance of learning a good latent embedding, we also explored the learning paradigm where we first optimize latent representation in an unsupervised fashion and then train the classifier using the already learnt embedding. To this end we trained an unsupervised PertVAE on all perturbation pairs and afterwards fitted Ridge LR classifiers, one using the PertVAE's latent representation of pre-treatment gene expression (PertVAE+LR on Z1) and another on the latent representation of predicted post-treatment state (PertVAE+LR on Z2). Additionally, we compared these results to the baseline models trained on principal component analysis (PCA) projection of the dataset to the first 100 principal components. The third best average performance over all models was achieved by LR trained on the latent embedding of pre-treatment gene expression learned by the PertVAE model. The improvement is only $2.58\%$ over Ridge LR and does not beat SSVAE or Dr.VAE on any of the 19 drugs. Ridge LR performs better on PertVAE latent representations than on both the observed gene expression and PCA representation with the same number of hidden units (principal components) as PertVAE.

As the evaluation metric we use the area under precision-recall curve (AUPR) and area under the ROC curve (in the Supplementary Materials). Performance of all models is presented in Table \ref{tab:aupr}.

\begin{table}
  \caption{Cross-validated test AUPR (area under PR curve) of our Dr.VAE to SSVAE and other classification models. Methods including PCA and PertVAE are 2-step methods: (i) fit the unsupervised model, (ii) use latent representation to fit a standard classifier. The performance comparison is presented as the relative change to Ridge LR classifier trained on the pre-treatment gene expression.}
  \label{tab:aupr}
  \centering
  \includegraphics[width=1.\textwidth]{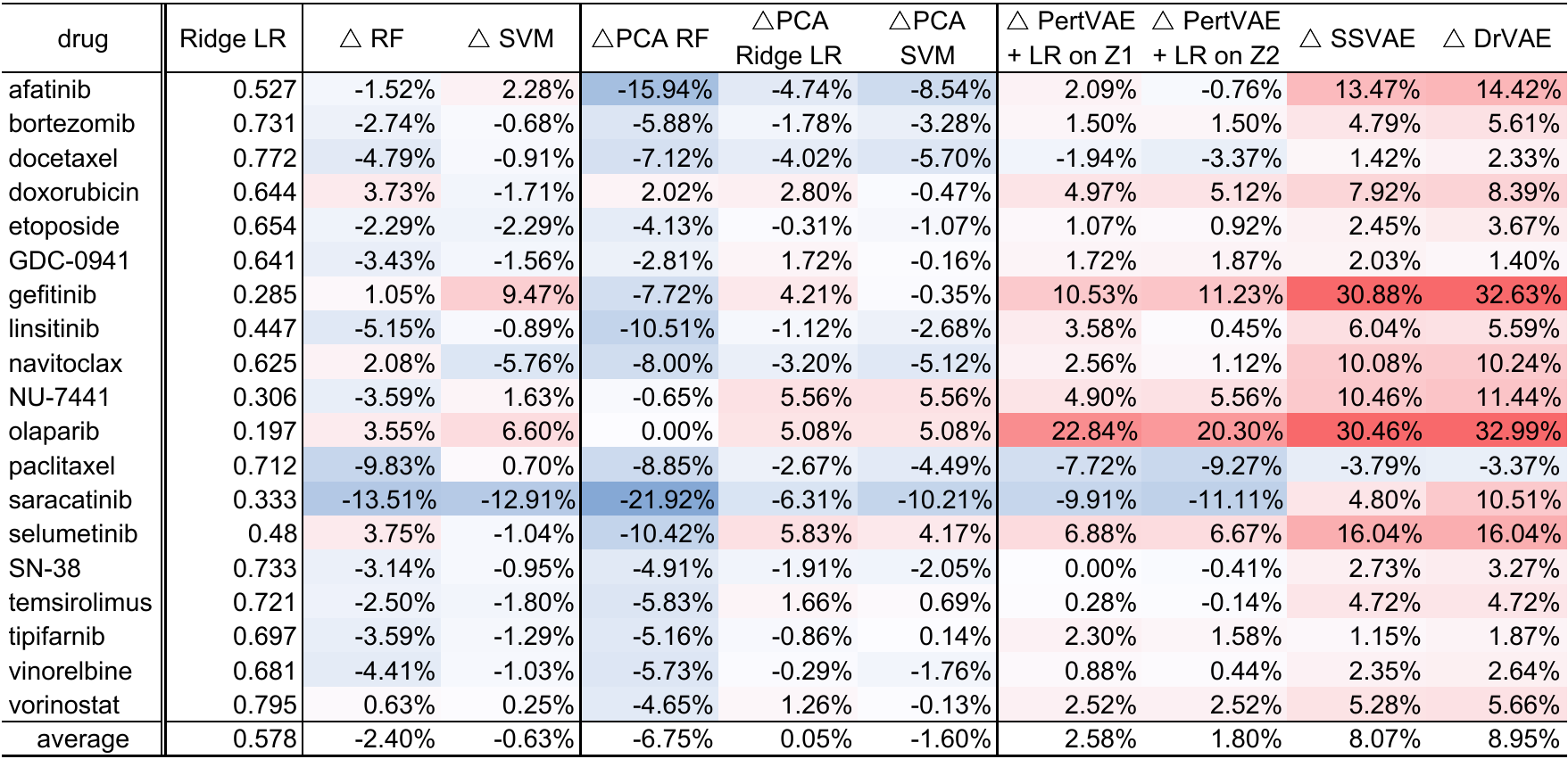}
\end{table}

\subsection{Reconstructing gene expression}
Variational Autoencoder is an expressive non-linear model, while PCA has the best reconstruction loss among linear models. To evaluate how well a VAE with our architecture can model gene expression, we fitted a VAE with various number of stochastic latent variables and compared its reconstruction to reconstructions by a PCA with equivalent number of principal components. As the measure of reconstruction quality we used Spearman's $\rho$ between reconstruction mean and the original gene expression. We plot the results in Figure \ref{fig:pert-reconst}. The Variational Autoencoder with our encoder/decoder architecture, as used in Dr.VAE and PertVAE, does better for small latent spaces ($<20$) after which it seems to overfit compared to PCA. We chose this architecture and 100 stochastic units as the default for all our models. We expect our models then to have enough expressive power and capacity to not just model gene expression but also find such a latent space that can be informative  for drug response and/or drug effect can be modeled as a stochastic linear function.

\subsection{Predicting post-treatment gene expression}
We trained a PertVAE for each drug independently to see how well we can predict drug perturbation effects. That is, we optimized the $\text{ELBO}_{\text{PertVAE}}$ and stopped training when perturbation prediction loss started to increase on the validation set.

To evaluate the prediction performance we computed Spearman's correlation $\rho_{\text{pred,pert}}$ between the mean of predicted gene expression distribution $\E_{\q(\z{1},\z{2}|\x{1})}[\p(\x{2} | \z{1})]$ and the observed post-treatment gene expression in the test set. We compare this correlation to the correlation $\rho_{\text{rec,pert}}$ between the mean of pre-treatment reconstruction distribution $\E_{\q(\z{1}|\x{1})}[\p(\x{1} | \z{1})]$ and the true post-treatment gene expression. This is done to assess whether the drug perturbation function is in fact learning anything beyond reconstructing pre-treatment gene expression. Note, that in the training step the ``drug effect'' mean function is initialized close to identity. If PertVAE would either underfit or overfit on the training set, we would expect $\rho_{\text{pred,pert}}$ to be no larger than $\rho_{\text{rec,pert}}$. Therefore we calculate Mann-Whitney single-sided test with the alternative hypothesis $\mathrm{H}_1 = \rho_{\text{rec,pert}} < \rho_{\text{pred,pert}}$ on the results of our 10-times randomized 5-fold CV. The average correlation values and p-values of the statistical test are in Table \ref{tab:pert-rho}, showing that PertVAE can at least partially predict drug perturbations for 5 out of 8 drugs (p-value $\leq 0.001$) for which the data set consists of perturbation experiments with at least 51 unique cell lines.

\begin{table}
\begin{minipage}[b]{.45\textwidth }
  \centering
  \includegraphics[width=1.\textwidth]{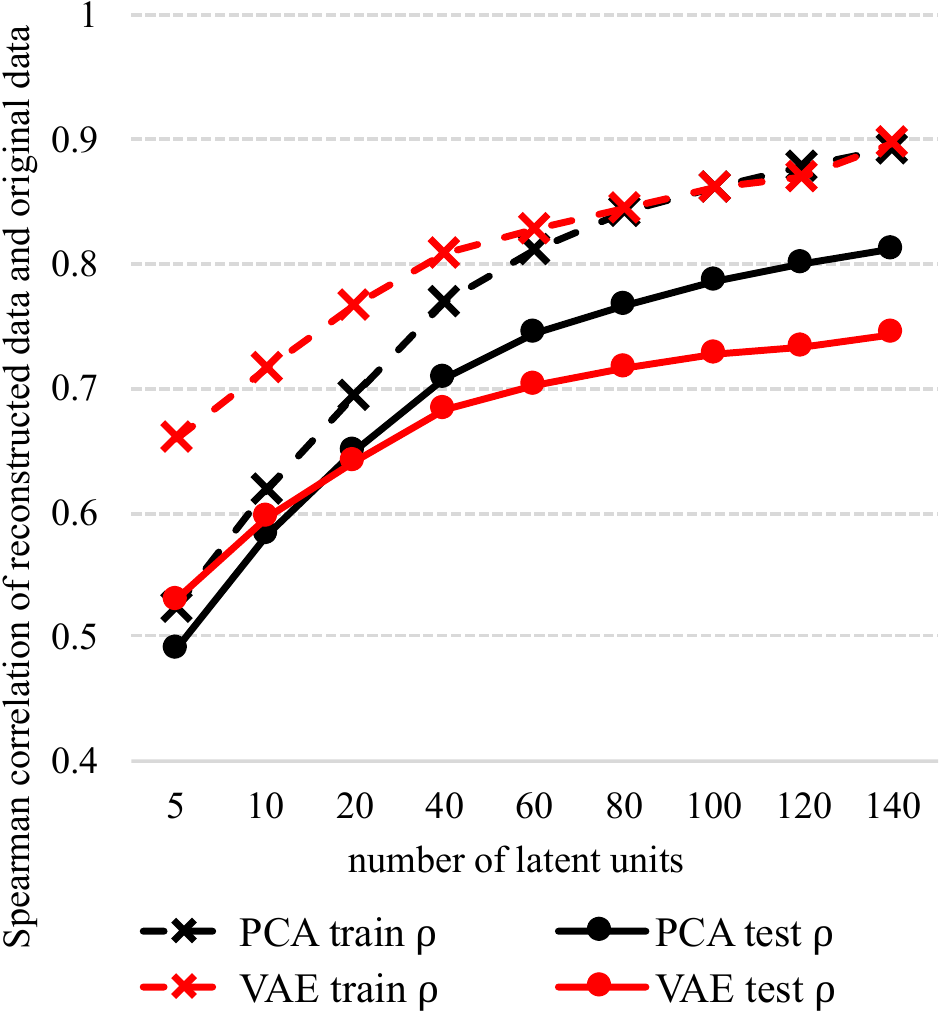}
  \captionof{figure}{PCA and VAE reconstruction quality comparison for varying latent space size.}
  \label{fig:pert-reconst}
\end{minipage}
\hspace{10pt}
\begin{minipage}[b]{.52\textwidth}
  \caption{Perturbation VAE prediction results with latent space size 100.}
  \label{tab:pert-rho}
  \centering
  \includegraphics[width=.95\textwidth]{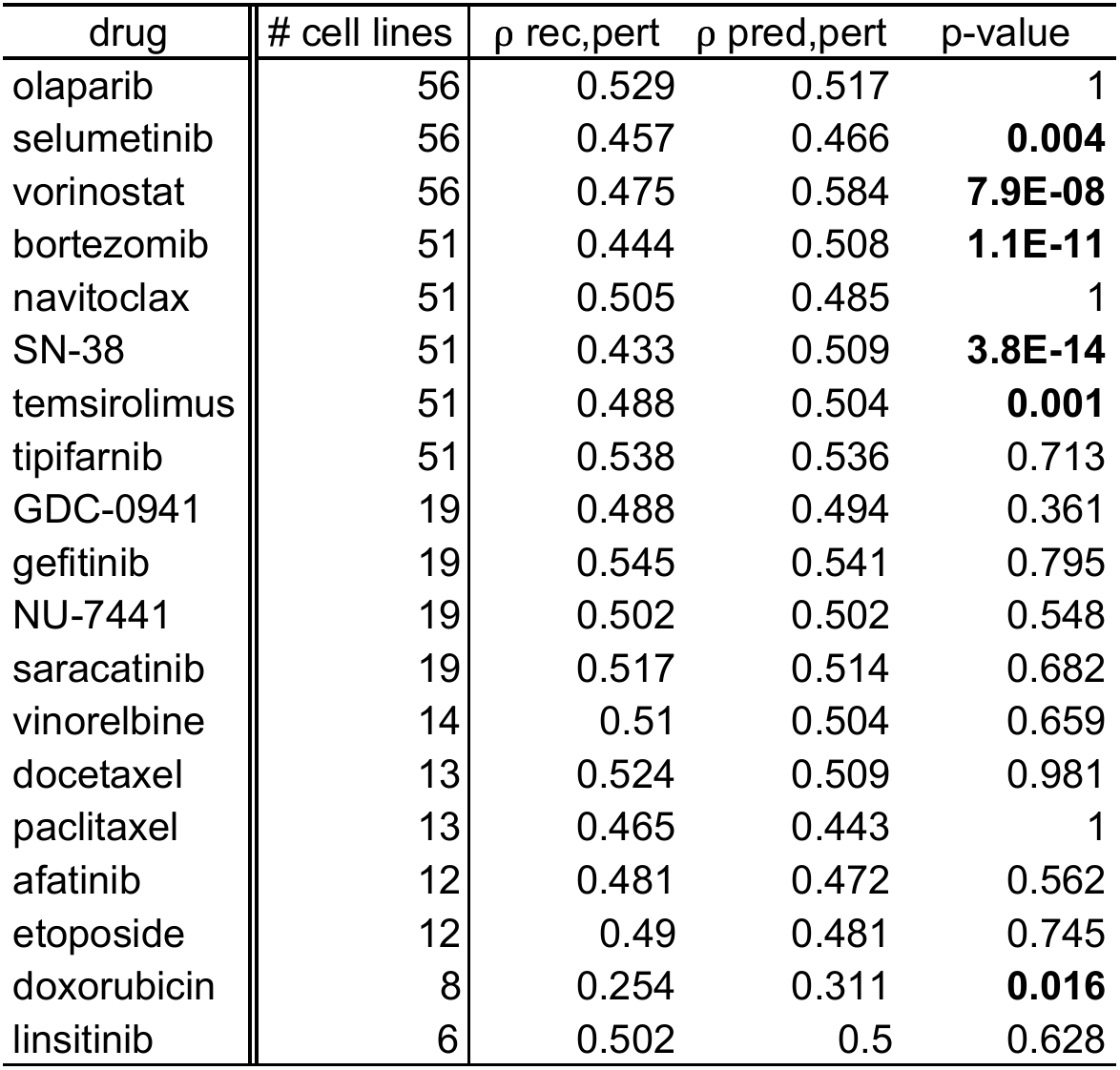}
\end{minipage}
\end{table}

\vspace*{-5pt}
\section{Discussion}
\vspace*{-5pt}
In our explorations of optimizing latent space, we found that doing well on reconstruction task does not directly lead to improved classification performance. The ability of Dr.VAE to model drug-induced perturbation effects on gene expression leads to limited improvement in classification performance. Compared to fitted PertVAE, a Dr.VAE model trained predominantly for classification does not learn to predict perturbation effects along the way. However, it is the best performing classification model. This set of observations compells us to conclude that the latent space serves a different role than simply compressing observed gene expression. Given a very small set of samples, very heterogeneous and noisy input and likely noisy output, the goal of the latent space is to capture the essense of the observed gene expression that is most useful and likely biased for prediction. The original goal of our work was to create a rich paradigm where much of the available data can be incorporated to boost the predictive performance of drug response. 
\comment{While the Bayesian paradigm lends itself to interpretation easier than more tradiational deep learning models (which btw did not work at all on our problem due to small sample size), our observations of reconstruction and perturbation predictions make us question whether given the currently available data, it will be easy to interpret the learned models.}
We did achieve an improvement in predicting drug response in the flexible and powerful VAE framework that we believe is the way to model such data in the future. 

\bibliographystyle{apalike} 
\renewcommand*{\bibfont}{\small}
\bibliography{refs}

\appendix

\clearpage
\section{Data sets summary}
\begin{table}[!ht]
  \caption{Dataset summarization. (a) Number of labeled and unlabeled pre-treatment singletons and perturbation pairs. In L1000 perturbation columns, the total number of experiments including biological replicates is shown, while the number of unique cell lines is listed in parentheses. (b) Shows split of the response-labeled samples to negative and positive classes. }
  \label{tab:data-num}
  \centering
  \subfigure[]{ \label{tab:data:a}
      \begin{minipage}[]{0.52\textwidth}
      \centering
      \includegraphics[width=1\textwidth]{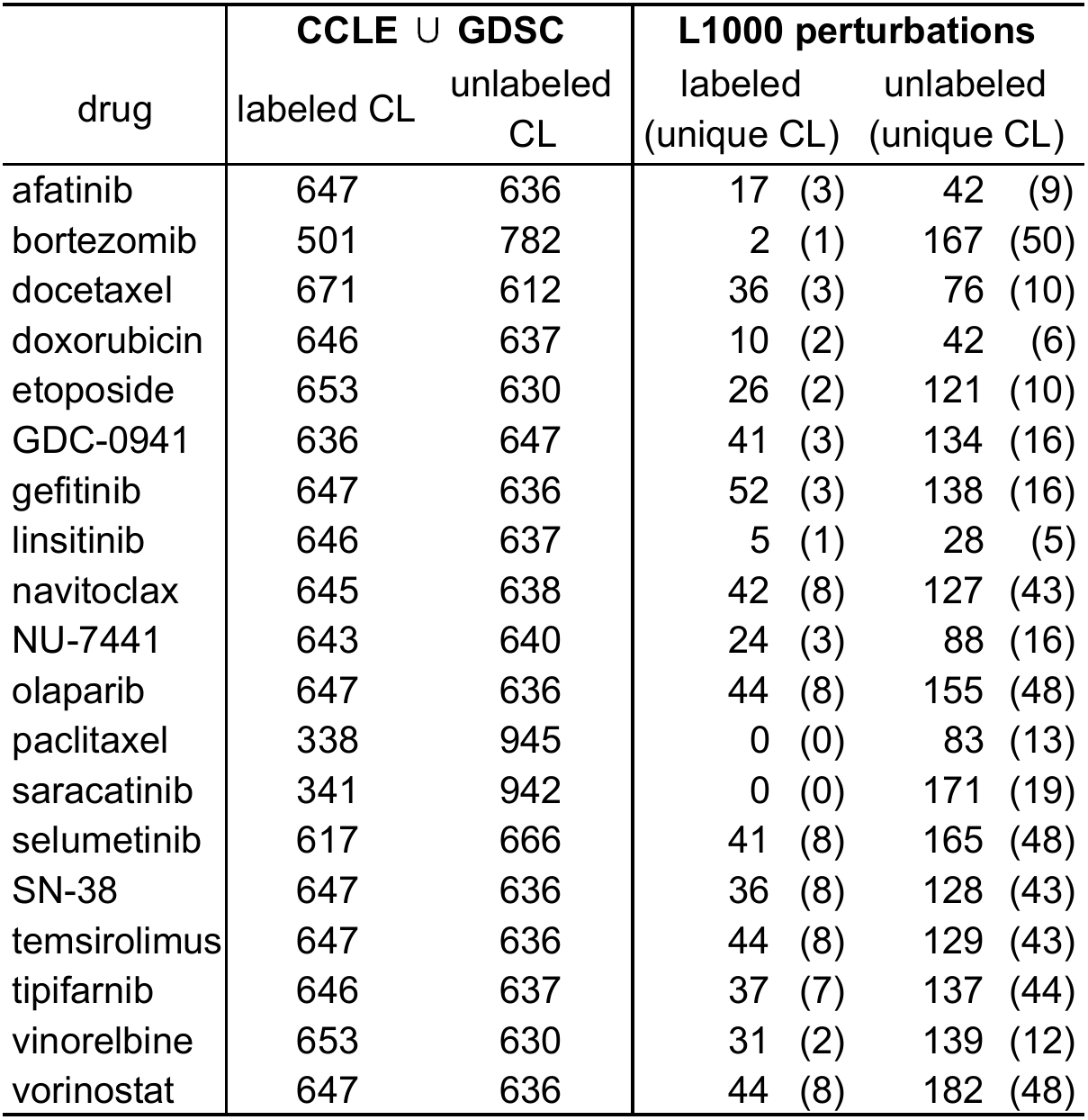}
      \end{minipage}
  }
\hspace{20pt}
  \subfigure[]{ \label{tab:data:b}
      \begin{minipage}[]{0.296\textwidth}
      \centering
      \vspace{10pt}
      \includegraphics[width=1\textwidth]{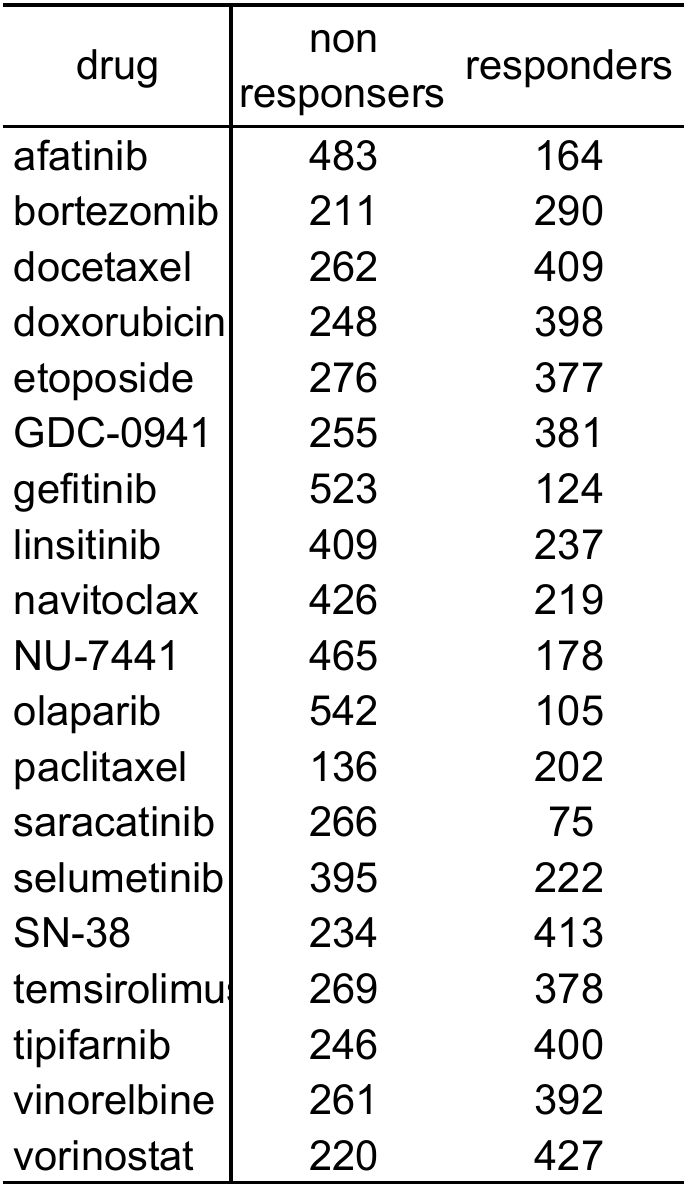}
      \end{minipage}
  }
\end{table}

\clearpage
\section{AUROC Results}
\begin{table}[!ht]
  \caption{Cross-validated test AUROC (area under ROC curve) of our Dr.VAE to SSVAE and other standard classification models. Methods including PCA and PertVAE are two step methods: (i) fit the unsupervised model and (ii) use latent representation to fit a standard classifier. Performance of the compared methods is presented as the relative change to Ridge logistic regression classifier trained directly on the pre-treatment gene expression.}
  \label{tab:auroc}
  \centering
  \includegraphics[width=1.\textwidth]{DrVAE_NIPS_aupr_v3-crop}
\end{table}

\end{document}